\documentclass[10pt,twocolumn,letterpaper]{article}

\usepackage{Style/cvpr}
\usepackage{times}
\usepackage{epsfig}
\usepackage{graphicx}
\usepackage{amsmath}
\usepackage{amssymb}
\usepackage{color}

\usepackage{mathtools}
\usepackage{enumitem}
\usepackage{subcaption}
\usepackage[export]{adjustbox}
\usepackage{booktabs}

\usepackage[pagebackref=true,breaklinks=true,letterpaper=true,colorlinks,bookmarks=false]{hyperref}

\cvprfinalcopy 

\def\cvprPaperID{1860} 

\ifcvprfinal\fi
\begin{document}

\title{Fine-Grained Classification of Pedestrians in Video: \\ Benchmark and State of the Art}

\author{David Hall and Pietro Perona\\
California Institute of Technology\\
{\tt\small {dhall,perona}@vision.caltech.edu}
}

\maketitle
\global\csname @topnum\endcsname 0
\global\csname @botnum\endcsname 0

\begin{abstract}
A video dataset that is designed to study fine-grained categorisation of pedestrians is introduced. Pedestrians were recorded ``in-the-wild'' from a moving vehicle. Annotations include bounding boxes, tracks, 14 keypoints with occlusion information and the fine-grained categories of age (5 classes), sex (2 classes), weight (3 classes) and clothing style (4 classes). There are a total of 27,454 bounding box and pose labels across 4222 tracks. This dataset is designed to train and test algorithms for fine-grained categorisation of people; it is also useful for benchmarking tracking, detection and pose estimation of pedestrians. State-of-the-art algorithms for fine-grained classification and pose estimation were tested using the dataset and the results are reported as a useful performance baseline.
\end{abstract}
\section{Introduction}
\label{sec-intro}
\begin{figure}[t]
\centering
    \includegraphics[width=0.19\linewidth,height=2.5cm]{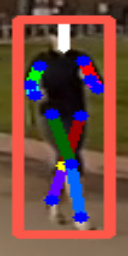}
    \includegraphics[width=0.19\linewidth,height=2.5cm]{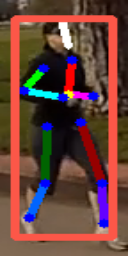}
    \includegraphics[width=0.19\linewidth,height=2.5cm]{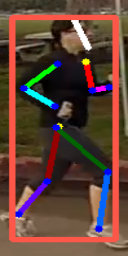}
    \includegraphics[width=0.19\linewidth,height=2.5cm]{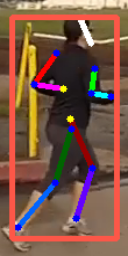}
    \includegraphics[width=0.19\linewidth,height=2.5cm]{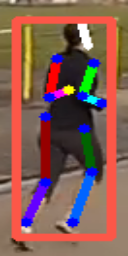}
    \includegraphics[width=0.19\linewidth,height=2.5cm]{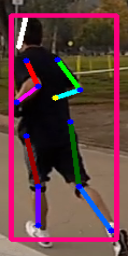}
    \includegraphics[width=0.19\linewidth,height=2.5cm]{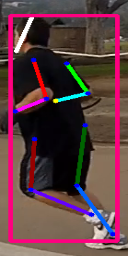}
    \includegraphics[width=0.19\linewidth,height=2.5cm]{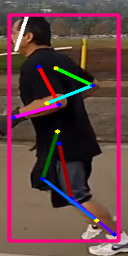}
    \includegraphics[width=0.19\linewidth,height=2.5cm]{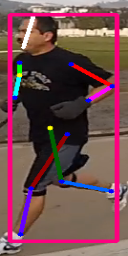}
    \includegraphics[width=0.19\linewidth,height=2.5cm]{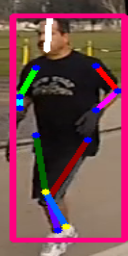}
    \includegraphics[width=0.19\linewidth,height=2.5cm]{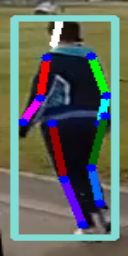}
    \includegraphics[width=0.19\linewidth,height=2.5cm]{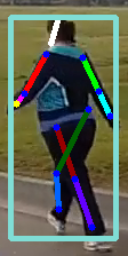}
    \includegraphics[width=0.19\linewidth,height=2.5cm]{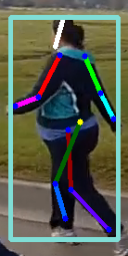}
    \includegraphics[width=0.19\linewidth,height=2.5cm]{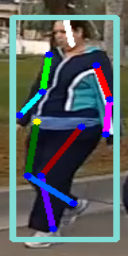}
    \includegraphics[width=0.19\linewidth,height=2.5cm]{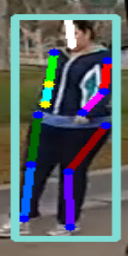}
    \caption{\textbf{Three examples from the CRP dataset.} Annotations include a bounding box, tracks, parts, occlusion, sex, age, weight and clothing style.}
    \label{fig:dataset-example}
\end{figure}
People are an important component of a machine's environment. Detecting, tracking, and recognising people, interpreting their behaviour and interacting with them is a valuable capability for machines. Using vision to estimate human attributes such as: age, sex, activity, social status, health, pose and motion patterns is useful for interpreting and predicting behaviour. This motivates our interest in fine-grained categorisation of people.

Visual classification involves recognising basic categories (e.g. `birds' vs. `chairs') and fine-grained categories, also called subcategories (e.g. `barn swallow' vs. `marten')~\cite{Branson2014a}. Since subcategories are similar in appearance subtle differences are often crucial. This is in contrast to basic categorisation where categories are visually distinct and therefore broad statistics of the image are often sufficient. 

While research in broad categorisation is well-supported by large datasets comprising thousands of categories~\cite{Deng2009,Lin2014}, fine-grained categorisation has so far been explored in a small number of domains including: animal breeds and species~\cite{Branson2010,Khosla2011,Wang2009}, plant species~\cite{Nilsback2006}, objects~\cite{Stark2012, Maji2013} and, what will be the focus of this work, people~\cite{Cao2008, Collins2009, Bourdev2011, Zhang2014}.  The availability of good-quality, large annotated datasets covering as many domains as possible is crucial for progress in fine-grained categorisation.

Prior work on the fine-grained categorisation\footnote{In the literature, fine-grained categorisation is more commonly referred to as attribute recognition in the human domain.} of people has typically focused on \textit{faces} with subcategories including: identity, age, sex, clothing type, facial hair and skin colour~\cite{Cottrell1990, Golomb1990, Moghaddam2002, Shakhnarovich2002, Baluja2006, Kumar2008}.

Fine-grained classification using the entire human body is still a relatively unexplored area. The current benchmark using images is ``The Attributes of People Dataset''~\cite{Bourdev2011} which was introduced in 2011. It includes 9 subcategories and has large variations in viewpoint and occlusion.  This dataset has been useful for researchers working on human attribute recognition~\cite{Bourdev2011,Zhang2014} but is limited by: a) its size, particularly when training deep networks~\cite{Zhang2014}; b) only bounding box annotations are provided and c) all subcategories are binary. 

There are also a number of video-based benchmark datasets~\cite{Sarkar2005, Iwama2012, Makihara2012} that are geared towards gait recognition. These datasets are limited by: a) the raw video footage is difficult to obtain with only silhouettes being readily available; b) subjects are cooperative (they know they are being filmed); c) the background is static and uncluttered; d) viewpoints are all profile; and e) subcategories are limited to identity and sex. 

\textbf{In this work, we introduce a public video dataset---Caltech Roadside Pedestrians (CRP)---to further advance the state-of-the-art in fine-grained categorisation of people using the entire human body.} 
Its novel and distinctive features are: 
\begin{enumerate}[nolistsep]
    \item Size (27,454 bounding box and pose labels) -- making it suitable for training deep-networks.
    \item Natural behaviour -- subjects are unaware, and behave naturally.
    \item Viewpoint -- Pedestrians are viewed from front, profile, back and everything in between.
    \item Moving camera -- More general and challenging than surveillance video with static background.
    \item Realism -- There is a variety of outdoor background and lighting conditions; examples can be found in Figure~\ref{fig-class_examples}.
    \item Multi-class subcategories -- age, clothing style and body shape.
    \item Detailed annotation -- bounding boxes, tracks and 14 pose keypoints with occlusion information; examples can be found in Figure~\ref{fig:dataset-example}. Each bounding box is also labelled with the fine-grained categories of age (5 classes), sex (2 classes), weight (3 classes) and clothing type (4 classes).
    \item Availability -- All videos and annotations are publicly available\footnote{\url{http://vision.caltech.edu/~dhall/projects/CRP}}
\end{enumerate}

\section{Related work}
\label{sec-related-work}
\begin{table}[t]
    \centering
        \begin{tabular}{@{} l *2c @{}}    \toprule
            \textit{Number of Frames Sent to MTURK} & 38,708 \\ 
            \textit{Number of Frames with at least 1 Pedestrian} & 20,994 \\ 
            \textit{Number of Bounding Box Labels} & 32,457 \\ 
            \textit{Number of Pose Labels} & 27,454 \\ 
            \textit{Number of Tracks} & 4,222 \\ 
        \bottomrule
        \end{tabular}
    \caption{Dataset Statistics}
    \label{tab-stats}
\end{table}
\begin{figure}[t]
    \centering
        \includegraphics[width=0.47\linewidth, height=8.5cm]{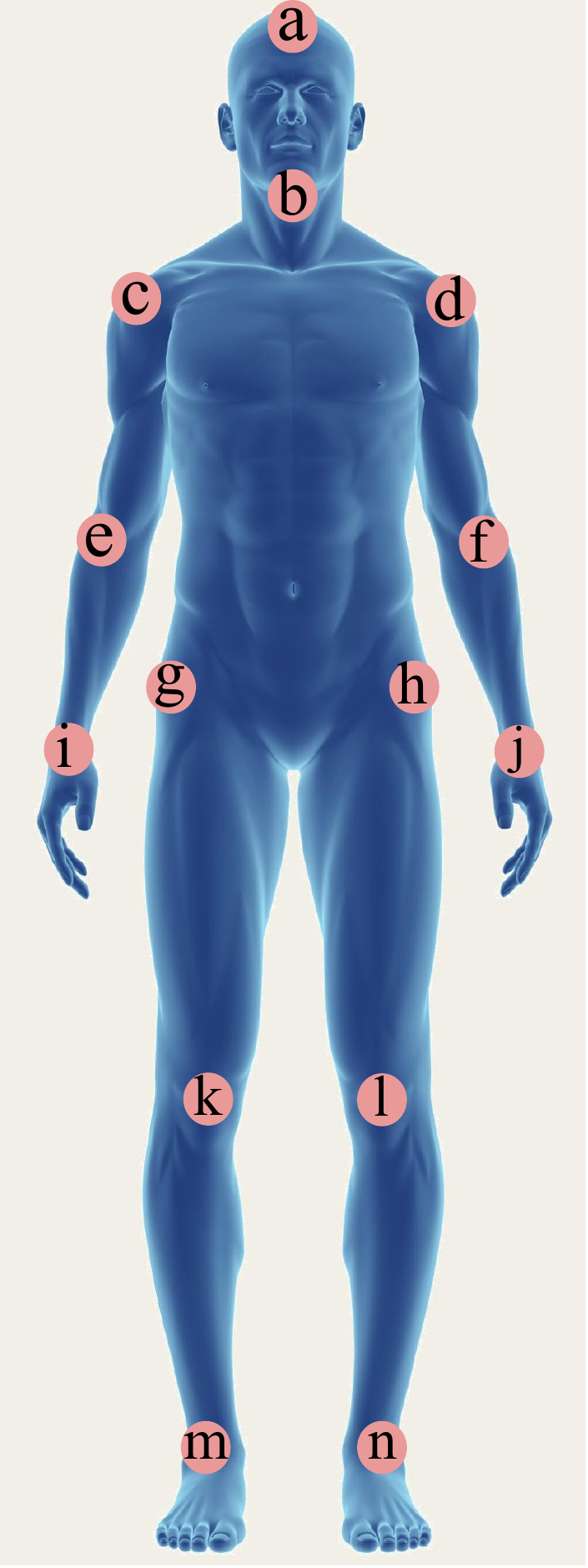} \quad
        \includegraphics[width=0.47\linewidth, height=8.5cm]{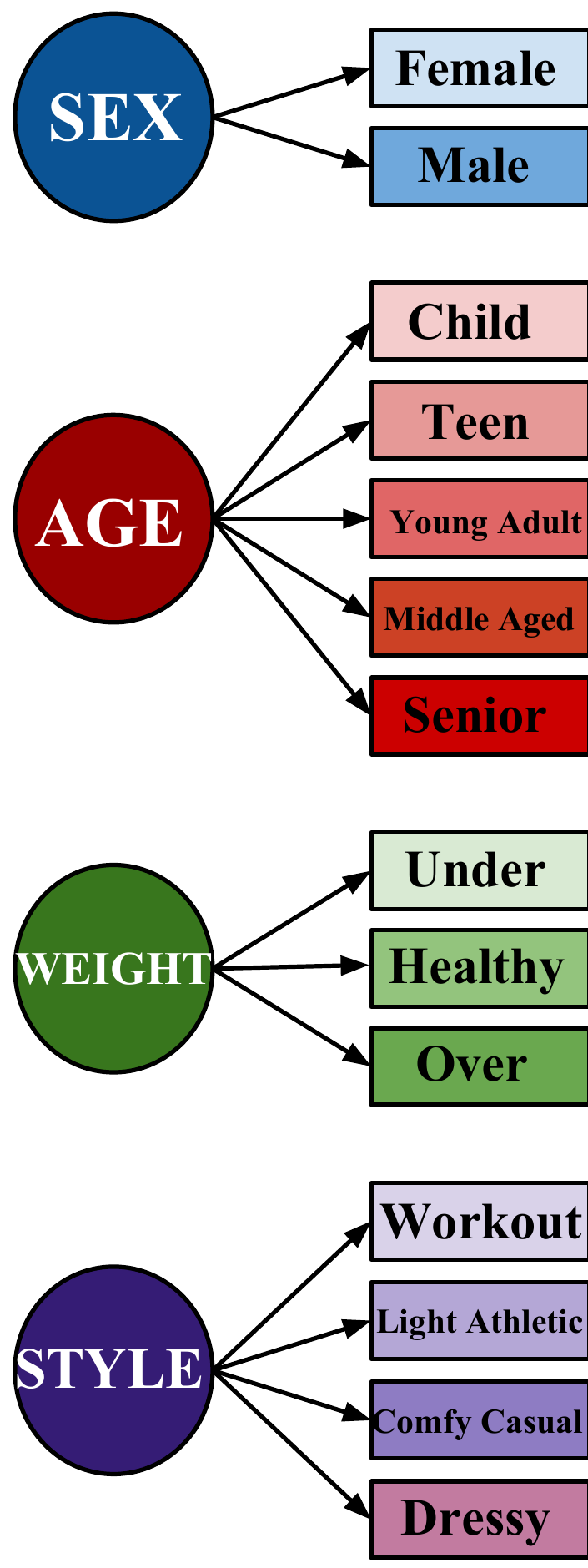}
        \caption{\textbf{(Left) The 14 body parts used as keypoints for pose annotation:} a) top of the head; b) chin; c) right shoulder; d) left shoulder; e) right elbow; f) left elbow; g) right hip; h) left hip; i) right wrist; j) left wrist; k) right knee; l) left knee; m) right ankle and n) left ankle. \textbf{(Right) The possible class labels for each of the four subcategories:} sex, age, weight and clothing style}
        \label{fig-keypoints}
\end{figure}
\begin{figure*}[t]
    \centering
        \includegraphics[width=0.30\linewidth]{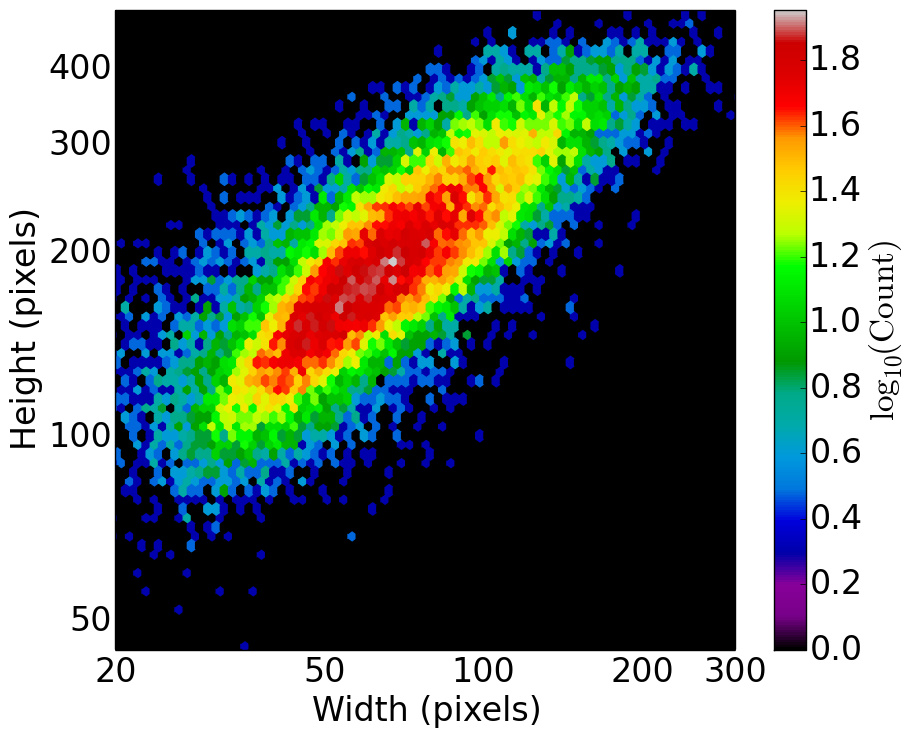} \qquad
        \includegraphics[width=0.30\linewidth]{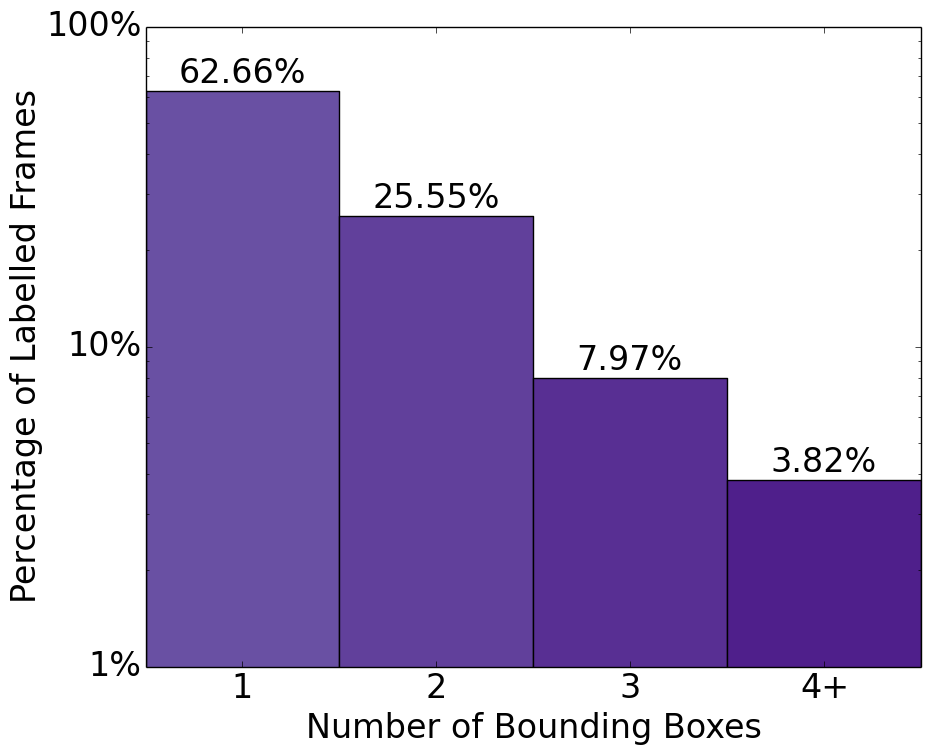} \qquad
        \includegraphics[width=0.30\linewidth]{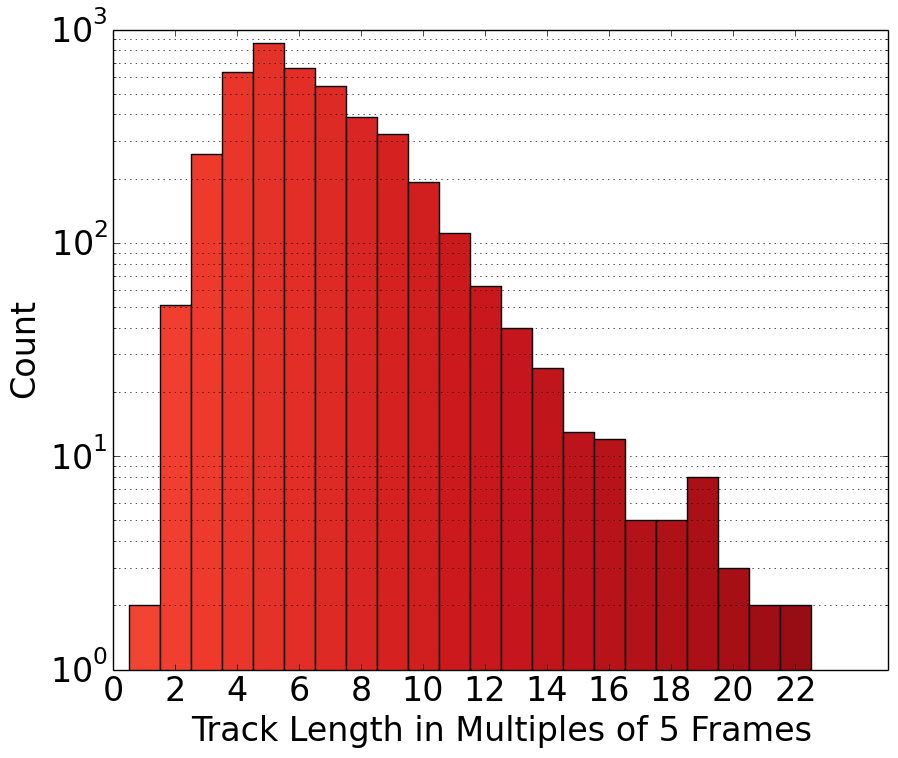}
        \caption{\textbf{(Left) A histogram of the height and width of the bounding boxes in the dataset.} The mean bounding box size is 71 pixels wide by 201 pixels high. The resolution is twice as large as Caltech Pedestrians~\cite{Dollar2009a} but 2.5 times smaller than the ``Attributes of People Dataset'', which is currently used for fine-grained classification~\cite{Bourdev2011}. \textbf{(Middle) A histogram of the number of bounding boxes in each of the labelled frames in the dataset.} 63\% of frames have only a single person in them, the remainder have two or more. \textbf{(Right) A histogram of the track length.} Since only every 5th frame is labelled, each pedestrian takes an average of 32.5 frames or 1.1 seconds to move through the field of view of the camera.}
        \label{fig-bbox_stats}
\end{figure*}
Existing fine-grained datasets on birds~\cite{Branson2010, Wah2011, Berg2014}, dogs~\cite{Khosla2011, Liu2012a, Parkhi2012}, cats~\cite{Parkhi2012}, butterflies~\cite{Wang2009}, flies~\cite{Mart2009}, leaves~\cite{Kumar2012}, flowers~\cite{Nilsback2006, Nilsback2008}, aircraft~\cite{Maji2013} and cars~\cite{Stark2012}, cover a single subcategory (usually species, breed or type) but have hundreds of classes. 

Fine-grained categorisation of people, however, began with a focus on the single \textit{binary} subcategory of sex. Using neural networks, SEXNET~\cite{Golomb1990} and EMPATH~\cite{Cottrell1990} were the first efforts to classify sex from faces; methods using support vector machines~\cite{Moghaddam2002} and boosting~\cite{Shakhnarovich2002, Baluja2006} soon followed. Work on classifying age and race from faces can also be found in the literature, with in-depth surveys available~\cite{Fu2010, Fu2014}. The first attempt at collecting a face dataset with multiple, multi-class subcategories was FaceTracer~\cite{Kumar2008}. It captured seven subcategories relevant to people, these included: sex, age (4 classes) and race (3 classes). Each subcategory had between 1000-4000 examples. 

In low resolution situations and particularly surveillance settings, faces are not suitable for fine-grained categorisation. This has led to work that looks at using the entire body, which presents additional cues such as clothing, body shape and motion patterns. Cao~\cite{Cao2008} took the existing MIT pedestrian dataset~\cite{Oren1997} and manually annotated it with sex labels; this was repeated by~\cite{Collins2009} who also labelled the VIPeR~\cite{Gray2008} dataset. It wasn't until ``The Attributes of People Dataset''~\cite{Bourdev2011} was released in 2011, that a full-body dataset, with more than a single subcategory, was publicly available. This dataset has 9 binary subcategories across 8035 images, with large variations in viewpoint and occlusion. Bounding box annotations are also provided. The dataset has a high resolution with an average bounding box size of 532 x 298 pixels and is the current benchmark. More recently, the ``Attributes 25K Dataset''~\cite{Zhang2014} was collected; it is a large dataset with 24,963 examples and the same subcategories as~\cite{Bourdev2011}, however, it is not publicly available.

The gait recognition community utilise the temporal information available in video for fine-grained categorisation, with a particular focus on identity. The task here is to infer the identity of someone from the way they walk. State-of-the-art methods for gait recognition extract a \textit{sequence} of silhouettes of a particular person. A Gait Energy Image~\cite{Han2006} or a Gait Entropy Image~\cite{Bashir2009} is then computed from the sequence of silhouettes once an estimate of the gait period is determined. These features are then used for classification. 

There are a number of video datasets available, the first being the USF HumanID Dataset~\cite{Sarkar2005}. It contains 1870 sequences of 122 individuals. It is collected outdoors with a static background. Participants are cooperating subjects, aware of being filmed, who are asked to walk a predefined elliptical path, with only the back portion used ensuring the viewpoint is always profile. Silhouettes are available for immediate download while the entire video collection can take up to 3 months to obtain. A more recent example is the Large Population Dataset~\cite{Iwama2012}. It contains 4016 subjects who each occur in 2 sequences. It is collected indoors with a green screen background. Participants walk a predefined path, however, the viewpoint varies from nearly-frontal to profile. Sex and age labels were also collected however these have not yet been released. Only silhouettes are available for download but only after authorisation is granted by the authors.

\begin{figure*}[t]
    \centering
        \includegraphics[width=0.24\linewidth, height=6.0cm]{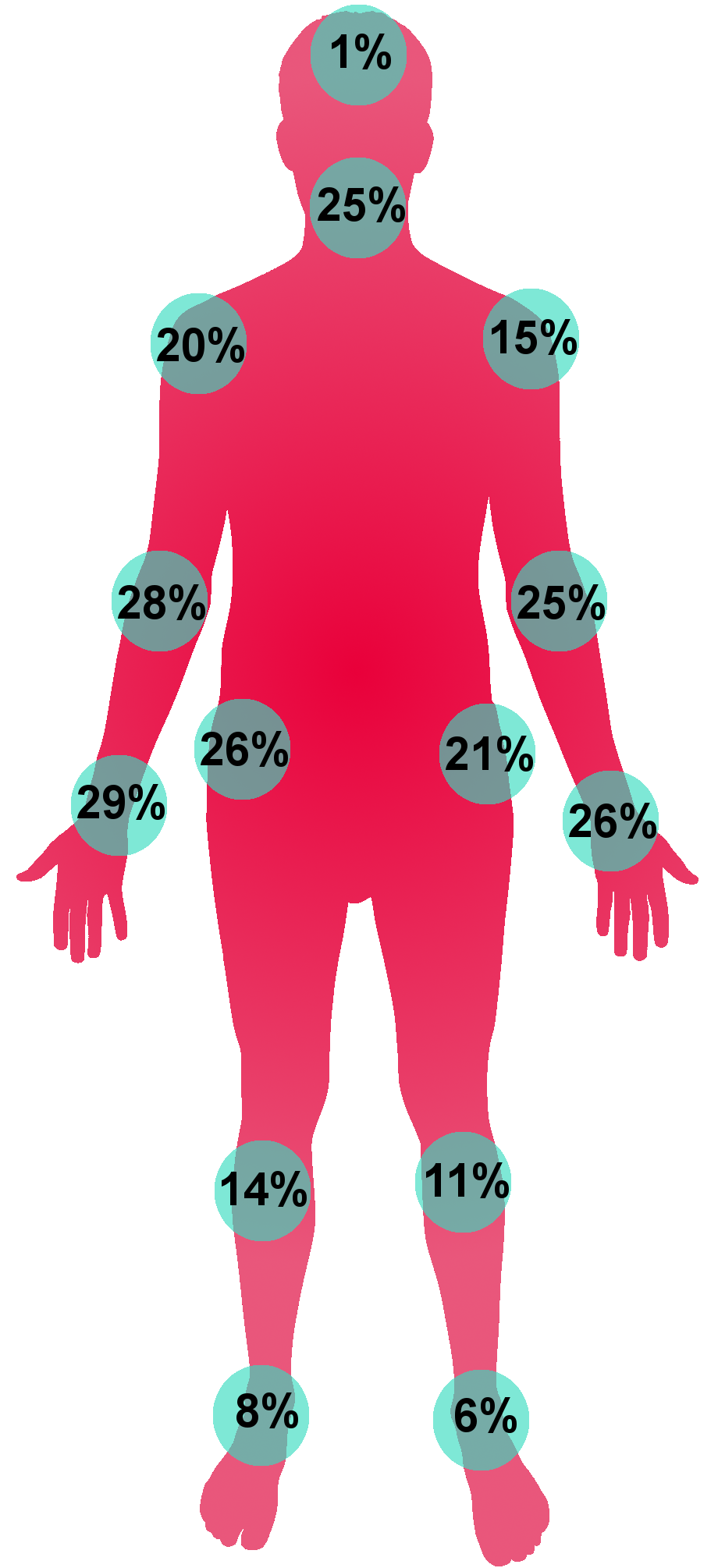} \qquad \qquad
        \includegraphics[width=0.60\linewidth, height=6.0cm]{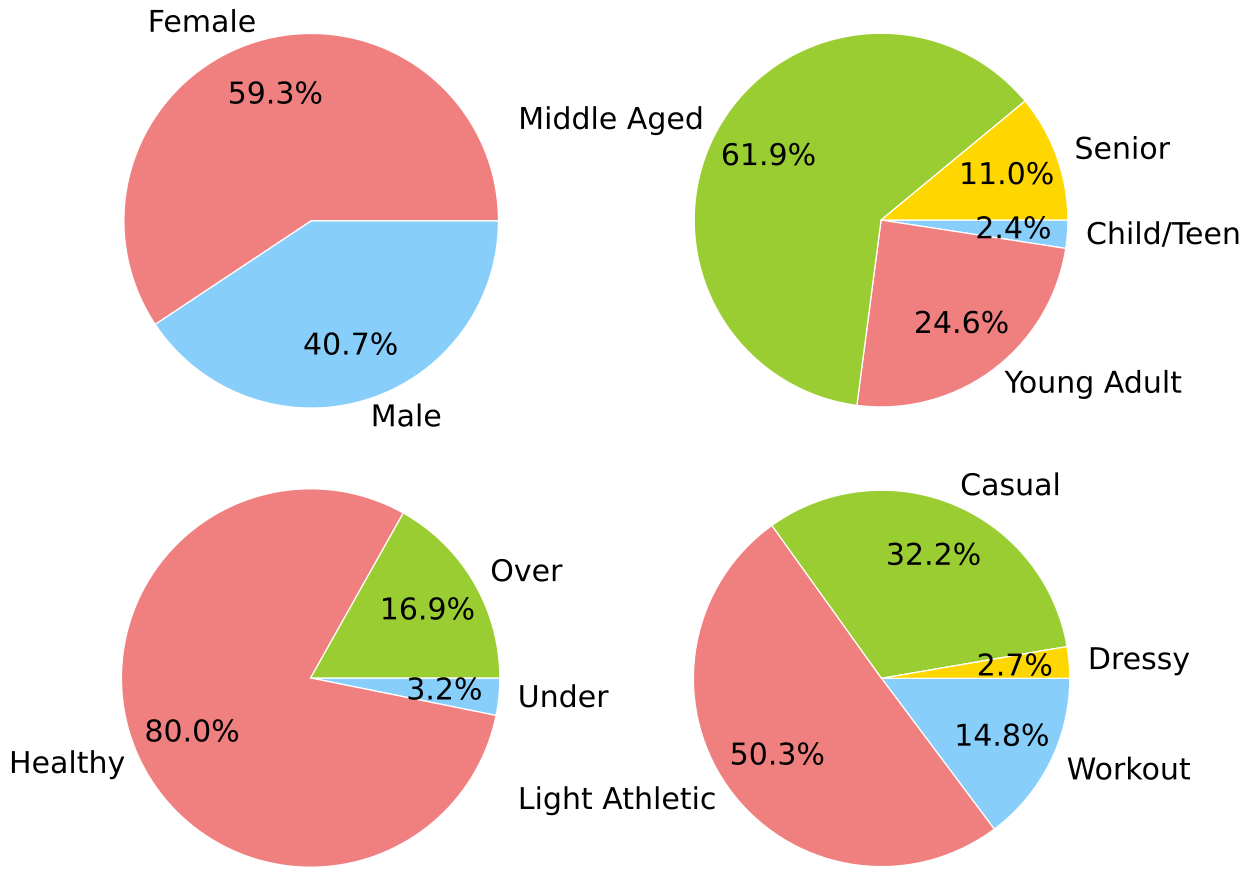}
        \caption{\textbf{(Left)} \textbf{The percentage of labels that are occluded for each keypoint.} The top of the head is rarely occluded, followed by the left and right ankles. The wrists, shoulders, elbows, hips and chin are all occluded around 25\% of the time. \textbf{(Right)} \textbf{The percentage of labels for each class} in the four fine-grained categories of the dataset. For all subcategories there is quite a large class imbalance. Very few children, teenagers, under weight or well-dressed people were seen. Given the setting, this is not surprising. The remaining classes have a reasonable number of labels.}
        \label{fig-attribute_stats}
\end{figure*}
\section{Dataset Collection}
\label{sec-dataset}
In this section we describe in detail, the method in which the videos of the dataset were collected and annotated. Due to the large number of annotations required it was important to develop an efficient and cost effective pipeline. For this reason, crowdsourcing, using workers from Amazon's Mechanical Turk (MTURK) was used for all of the annotation tasks. 

\subsection{Video Collection}
\label{sec-collection}
This dataset contains 7 videos. Each video is captured by mounting a rightwards-pointing, GoPro Hero3 camera to the roof of a car. The car then completes three laps of a ring road within a park where there are many walkers and joggers. The videos were shot using a wide-angle mode, at a resolution of 1280x720 pixels, and a frame rate of 30 fps. Each video has on average, 37,000 frames, for a total of 261,645 frames in the entire dataset. Each video was recorded at 8AM on different days of the week, over a 9 month period.  

\subsection{Bounding Box Annotation}
\label{sec-bb}
For each video, the first task was to annotate all of the pedestrians with bounding boxes. To make this a cost-effective task, a coarse-to-fine approach was used. Every 10th frame was sent to MTURK where three workers were instructed to draw a bounding box around every pedestrian in the image. The bounding boxes from each worker were then combined into a single set of bounding box labels for each frame using clustering. For this stage, a total of 26,168 frames were sent for annotation.

A further set of frames were sent for annotation so that every 5th frame of the video would be labelled. To avoid sending empty frames, the results from the coarse labelling attempt were used. For every frame $x$ that had a set of bounding box labels (the image actually contained pedestrians), two frames were sent to MTURK for labelling, frames $x+5$ and $x-5$. These frames were again annotated by three workers and a single set of bounding box labels were generated as before. For this stage, 12,540 frames were sent for annotation. A total of 32,457 bounding box annotations were collected.

\subsection{Track Annotation}
\label{sec-track}
The next task was to create tracks (the time trajectory of an individual) from the bounding boxes. To do this, a worker was given a cropped image of a person from frame $x$ (obtained using the bounding box labels). They were then instructed to make a selection from the set of cropped people from frame $x+5$ that matched the original image. There was also an option to select that there was no match. Every person with a bounding box over 100 pixels in height was labelled by three workers. The workers' annotations were combined using a majority vote. If there was disagreement between all three workers, the bounding box was assigned a no-match label. Tracks were formed by chaining together the bounding boxes until a no-match label was encountered. Tracks were then verified by an expert annotator. Their task was to eliminate short tracking gaps and to correct any other mistakes. A total of 4,222 tracks were collected.  
\begin{figure*}[t]
    \centering
    \includegraphics[width=0.48\linewidth, height=5.5cm]{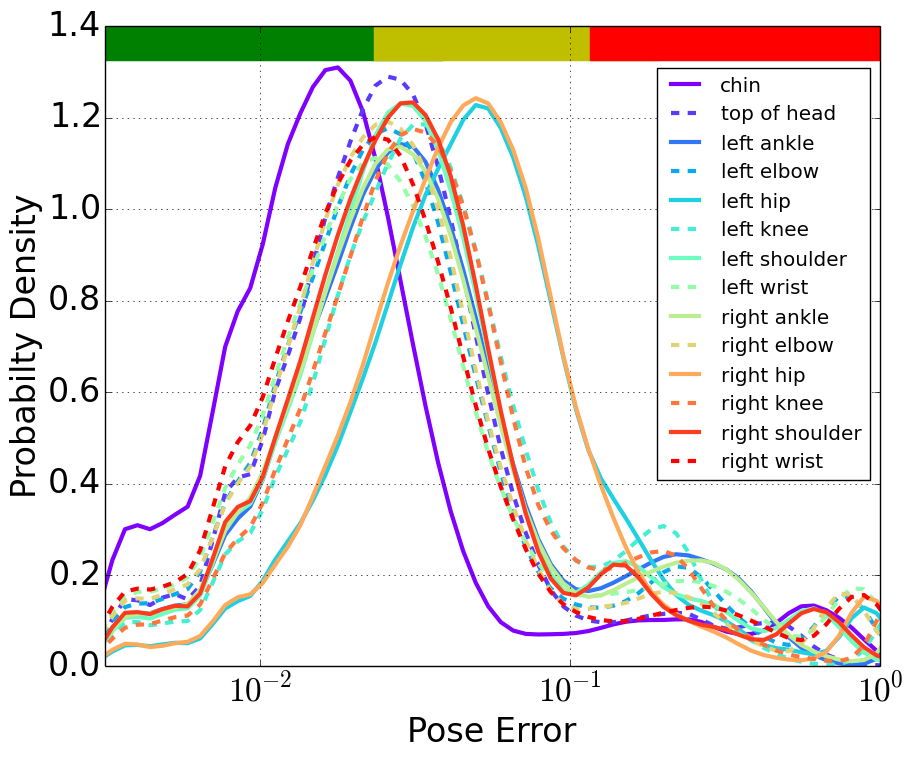} \quad
    \includegraphics[width=0.48\linewidth, height=5.5cm]{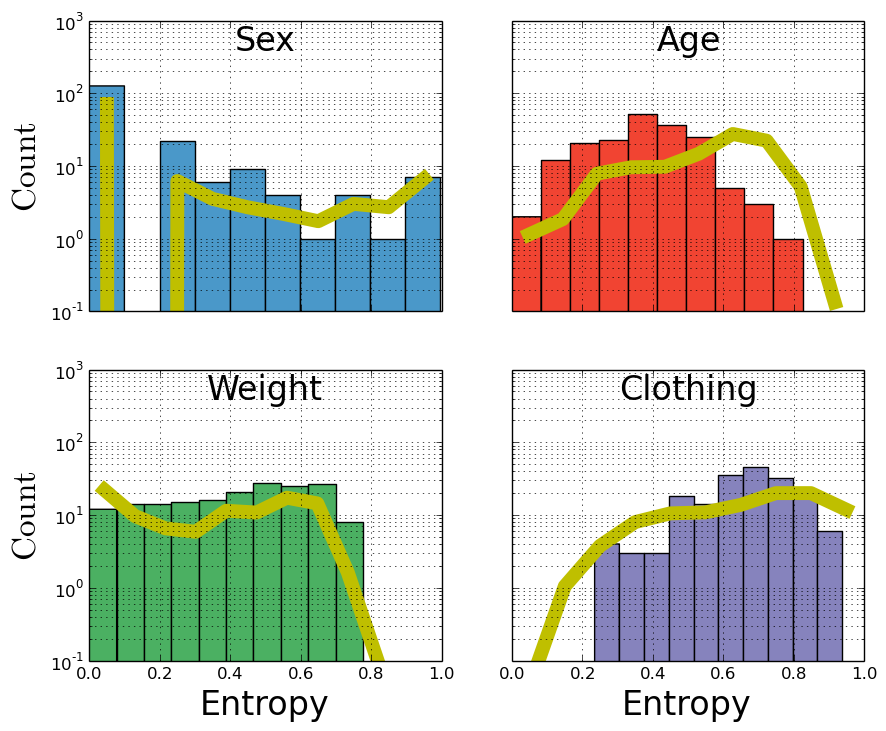}
    \caption{\textbf{(Left) An estimate of keypoint location error as a fraction of body height.} Given an image of a person, pose error is the distance between all of the workers keypoint locations, normalised by the height of the sample. The green band is where error is less than 3\%, the yellow band between 3-15\% and the red band is greater than 15\% (See Section~\ref{sec-pose_error}). \textbf{(Right) Estimates of the fine-grained labelling error for each subcategory.} The histograms correspond to the entropy of 30 workers labelling 180 randomly selected tracks from the dataset. The yellow line corresponds to our statistical model as described in Section~\ref{sec-fg_error}. The noise values in the model are 0.07, 0.15, 0.11 and 0.23 for sex, age, weight and clothing respectively. They need to be compared to the size of the bins, which depends on the number of classes in the subcategory. Bin sizes are equal to 0.5, 0.2, 0.33 and 0.25 respectively.}
        \label{fig-error}
\end{figure*}
\subsection{Pose Annotation}
\label{sec-pose}
To represent the pose of a human body, 14 body parts were used as keypoints which can found in Figure~\ref{fig-keypoints}. These are the same parts used in existing datasets~\cite{Johnson2010, Ramanan2006}. To annotate the parts, workers were given a cropped image of a person (obtained using the bounding box labels but with some extra padding) and instructed to click on one of the 14 body parts. If the part was occluded in any way, the workers were asked to right click where they thought the part was located. It was also possible to indicate if a part was not in the image. Every person with a bounding box over 100 pixels in height was labelled by three workers for each of the 14 keypoints. The workers annotations were combined by taking the median of the labels. A total of 27,454 pose annotations with 14 keypoints and occlusion information were collected. The pose annotations were then used to refine the bounding box labels since the worker labelled bounding boxes were not always tight. The refined bounding box is the tightest box that covers the set of keypoints.

\subsection{Fine-Grained Category Annotation}
\label{sec-attr}
\begin{figure*}[t]
    \centering
    \begin{tabular}{ccc}
        \includegraphics[width=0.32\linewidth, height=3.2cm]{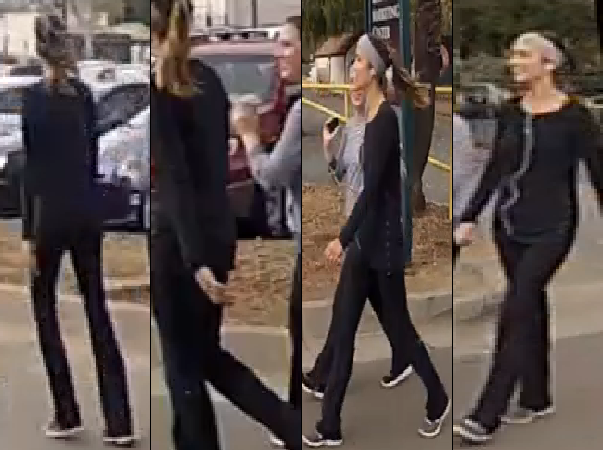}&
        \includegraphics[width=0.32\linewidth, height=3.2cm]{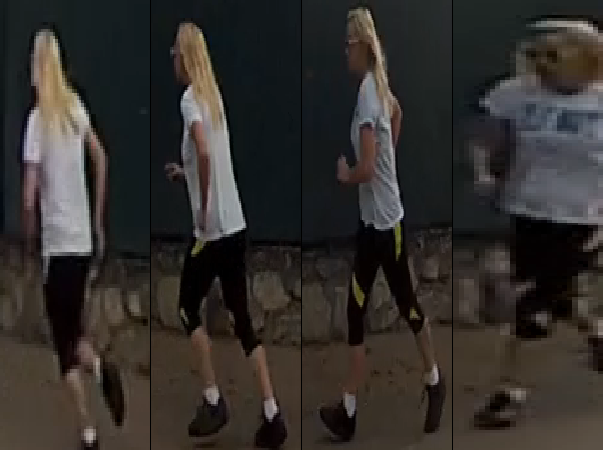}&
        \includegraphics[width=0.32\linewidth, height=3.2cm]{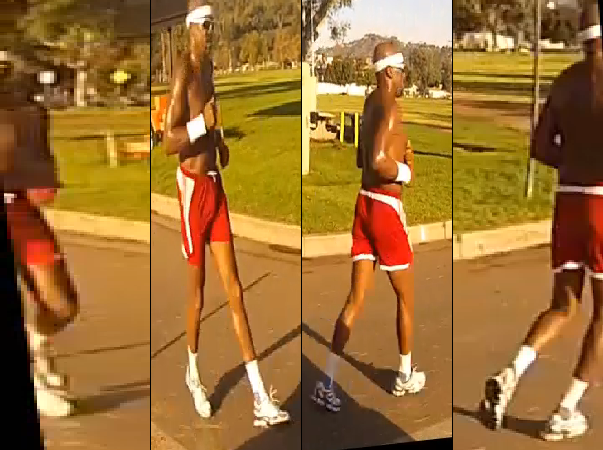}\\
        \includegraphics[width=0.32\linewidth, height=3.2cm]{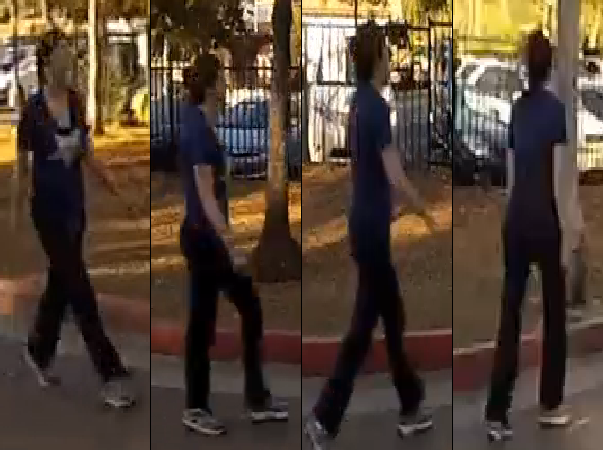}&
        \includegraphics[width=0.32\linewidth, height=3.2cm]{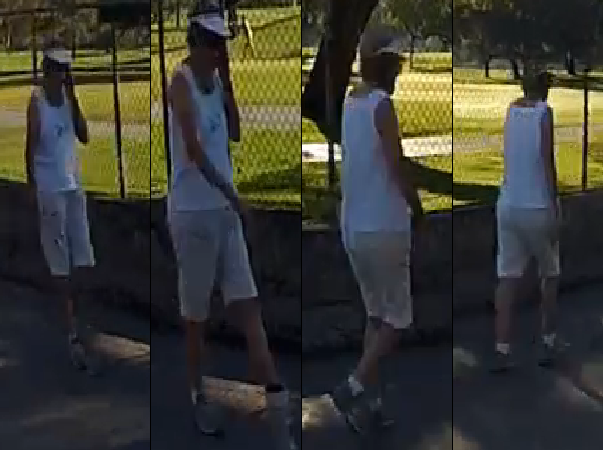}&
        \includegraphics[width=0.32\linewidth, height=3.2cm]{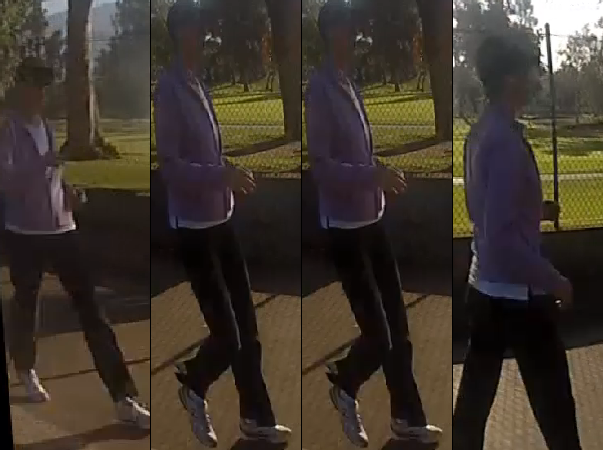}\\
        \includegraphics[width=0.32\linewidth, height=3.2cm]{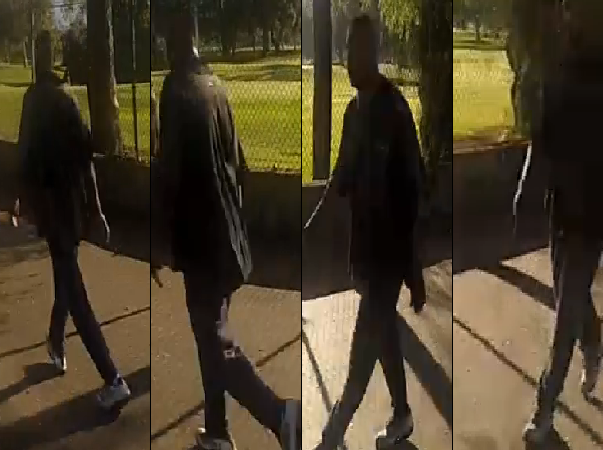}&
        \includegraphics[width=0.32\linewidth, height=3.2cm]{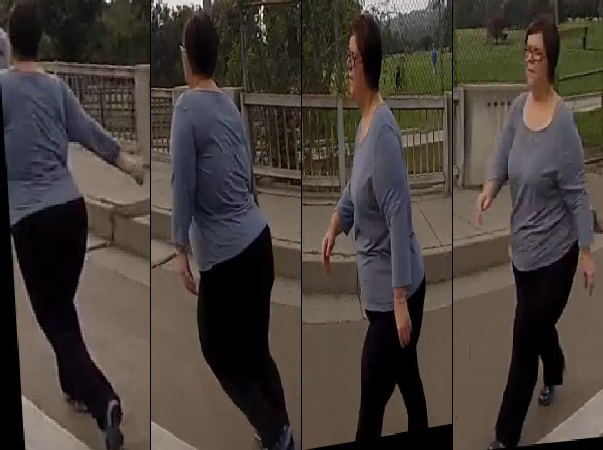}&
        \includegraphics[width=0.32\linewidth, height=3.2cm]{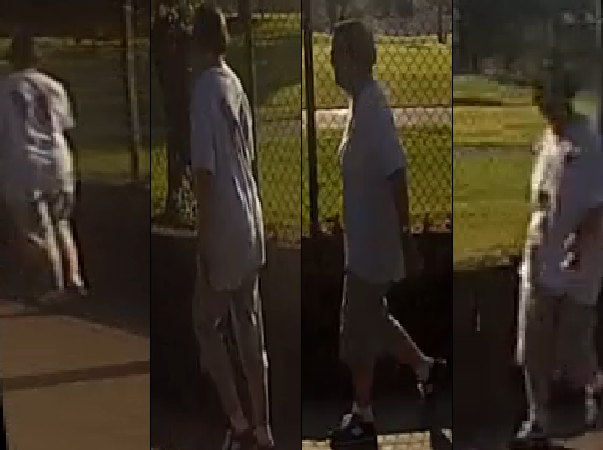}\\
        \includegraphics[width=0.32\linewidth, height=3.2cm]{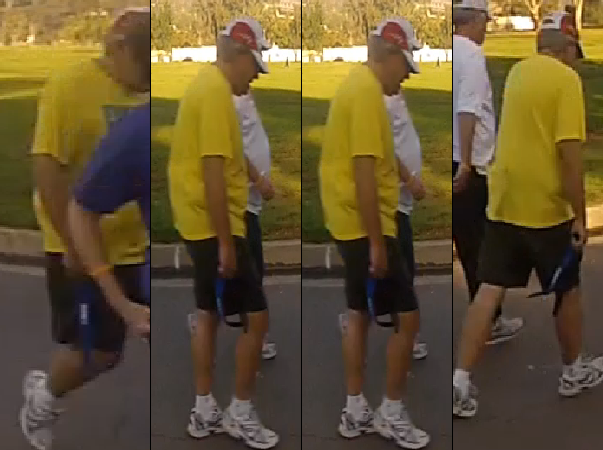}&
        \includegraphics[width=0.32\linewidth, height=3.2cm]{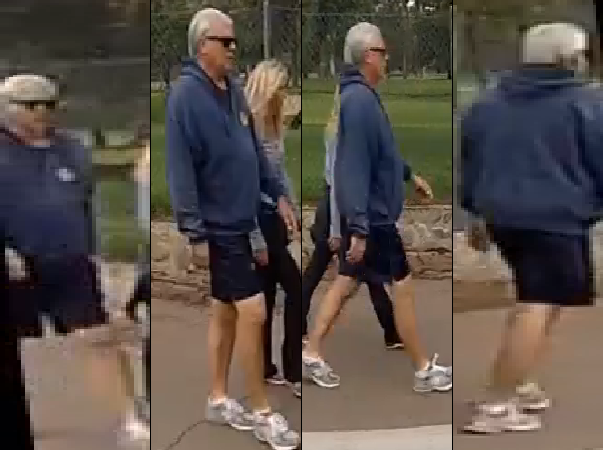}&
        \includegraphics[width=0.32\linewidth, height=3.2cm]{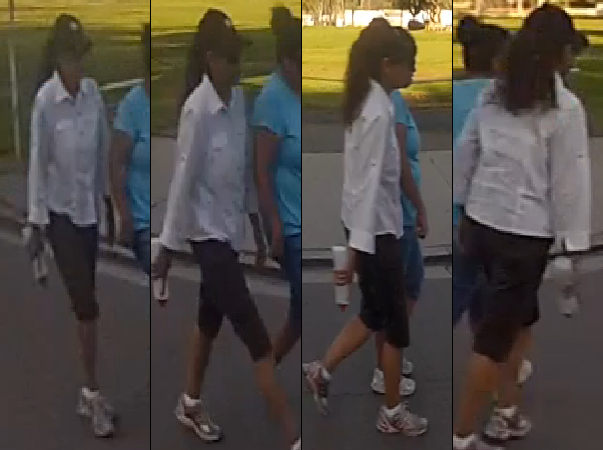}\\
        (A) & (B) & (C) 
        \end{tabular}
        \caption{\textbf{Class examples.} From top to bottom the corresponding class labels for each category are: (A) age - teen, young adult, middle aged, and senior; (B) weight - under, healthy, over, and over; (C) clothing style - workout, light athletic, casual comfort and dressy. The examples shown in this figure illustrate the variety of {\bf lighting conditions} (full sunlight, hazy, front and back lighting), {\bf viewpoints} (front, profile, back) and {\bf backgrounds}.}
        \label{fig-class_examples}
\end{figure*}

As mentioned in Section~\ref{sec-intro}, using vision to estimate human attributes is useful for interpreting and predicting behaviour. In this dataset we look at four fine-grained categories of people: sex, age, weight and clothing style. The possible class labels for these subcategories are shown in Figure~\ref{fig-keypoints}. While the classes for sex, age and weight were intuitive for us, those for clothing style were not. The 4 clothing style classes: workout, light athletic, comfortable casual and well-dressed (or dressy), were chosen in consultation with a fashion expert. The `workout class' groups people wearing spandex, singlet tops or no shirt at all. The `light athletic' class include people who are wearing yoga pants and tracksuits. The `comfortable casual' class contains people wearing shorts or items of clothing that would be typically worn in a casual setting. The `well-dressed' class are of people with button-up or collared shirts and dresses. Examples of these classes can be found in Figure~\ref{fig-class_examples}

To annotate the fine-grained categories, workers were given 4 examples of a person, sampled from one of the tracks. This allowed the worker to see the person from all possible viewpoints. They were then asked to select the best class label for one of the subcategories. For the clothing style task, workers were also shown examples of each class. The workers fine-grained labels were combined by taking a majority vote. If there was complete disagreement between workers a further 2 workers labelled the track and another majority vote was taken.    

\begin{figure*}[t]
    \centering
        \includegraphics[width=0.24\linewidth, height=2.6cm]{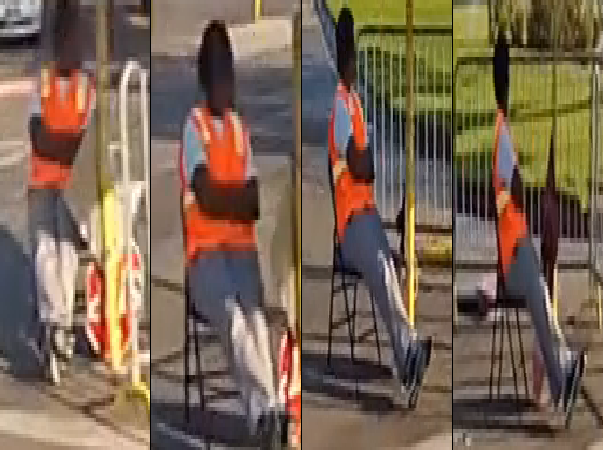}
        \includegraphics[width=0.24\linewidth, height=2.6cm]{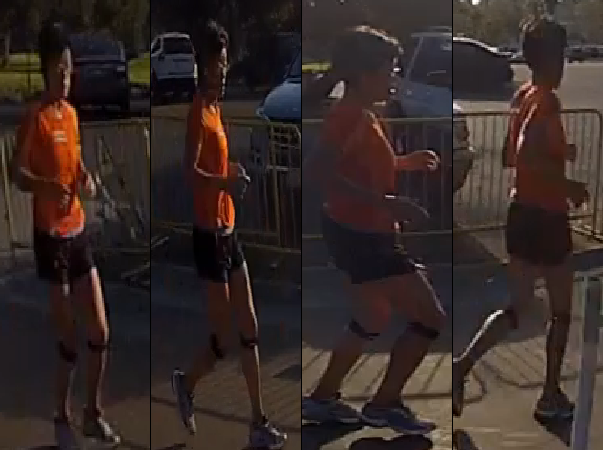}
        \includegraphics[width=0.24\linewidth, height=2.6cm]{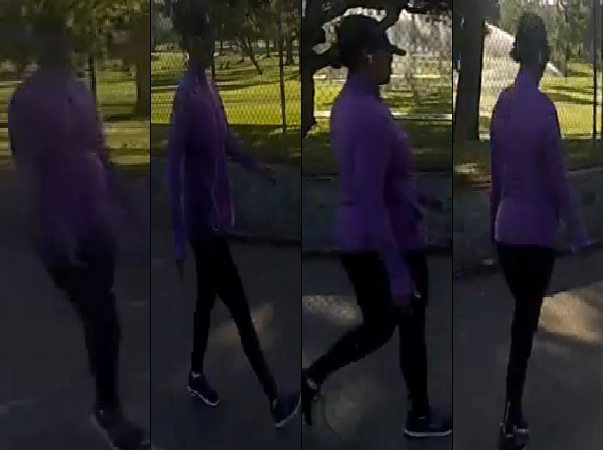}
        \includegraphics[width=0.24\linewidth, height=2.6cm]{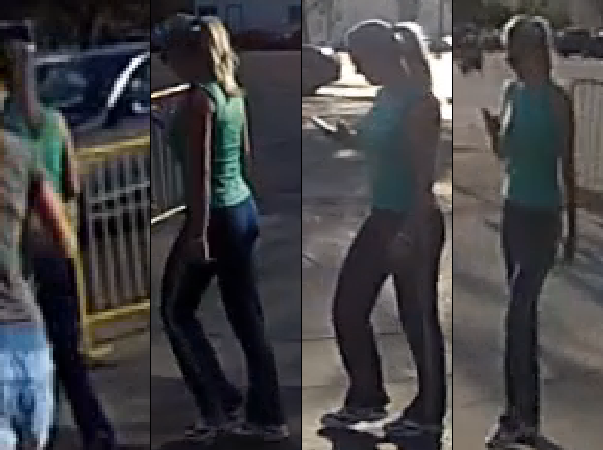}
        \includegraphics[width=0.24\linewidth, height=2.6cm]{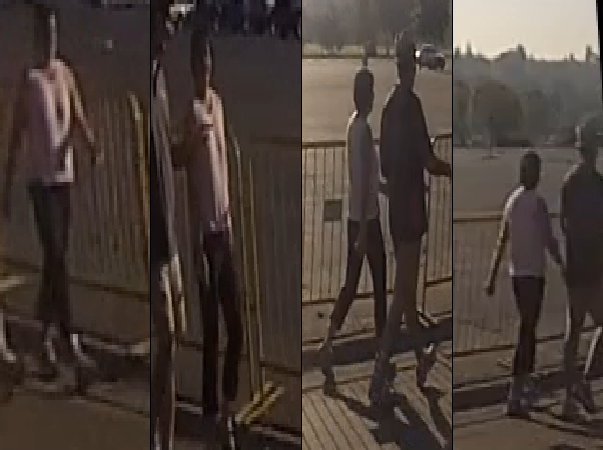}
        \includegraphics[width=0.24\linewidth, height=2.6cm]{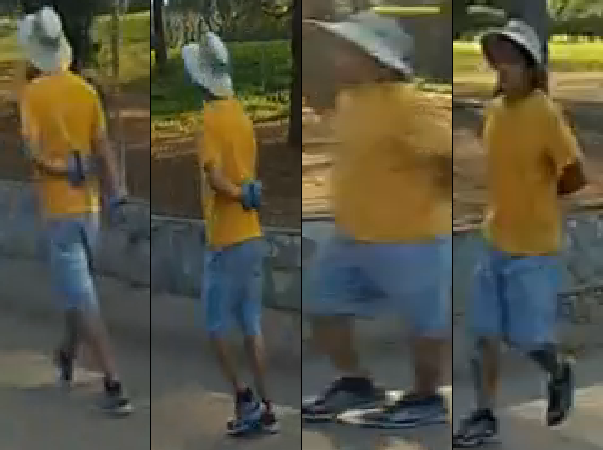}
        \includegraphics[width=0.24\linewidth, height=2.6cm]{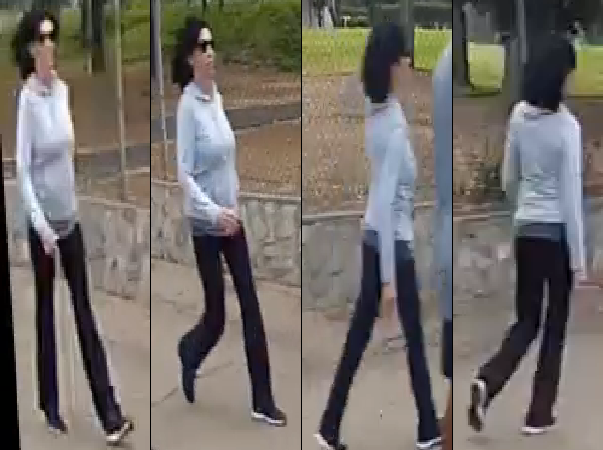}
        \includegraphics[width=0.24\linewidth, height=2.6cm]{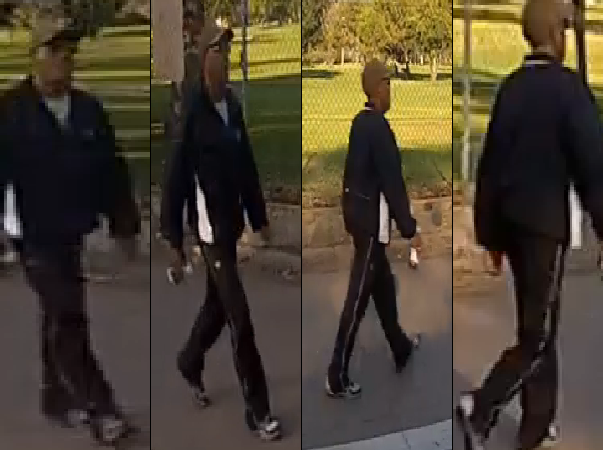}
        \includegraphics[width=0.24\linewidth, height=2.6cm]{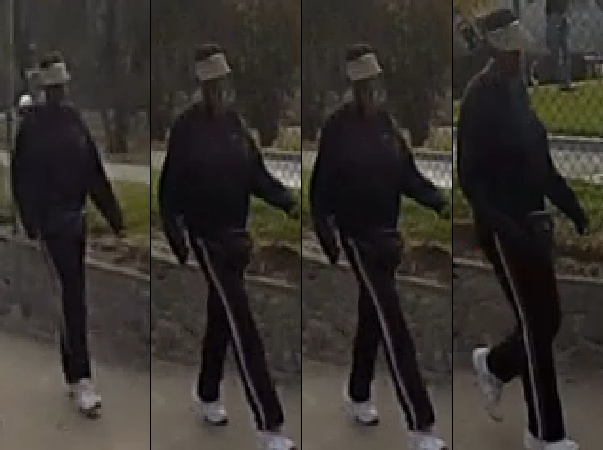}
        \includegraphics[width=0.24\linewidth, height=2.6cm]{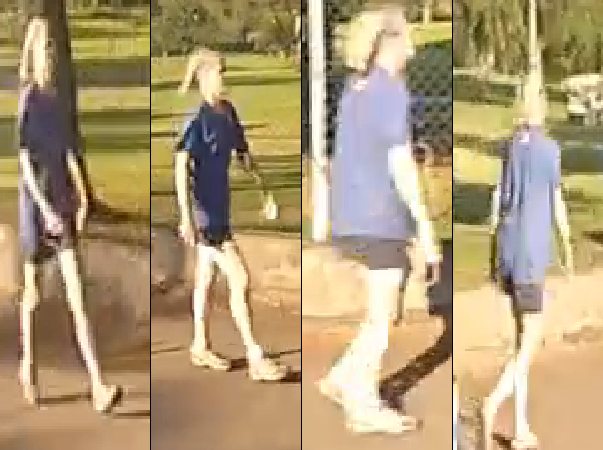}
        \includegraphics[width=0.24\linewidth, height=2.6cm]{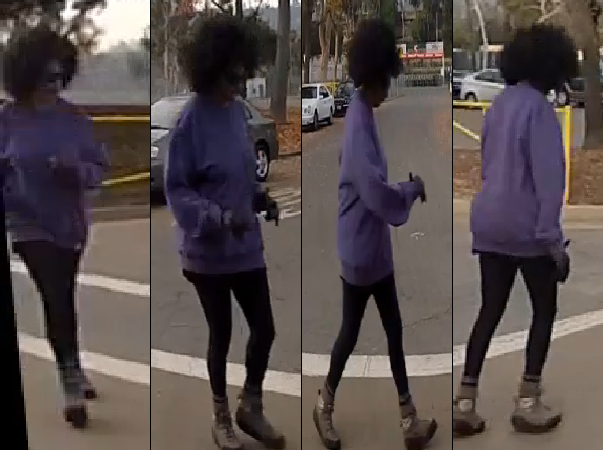}
        \includegraphics[width=0.24\linewidth, height=2.6cm]{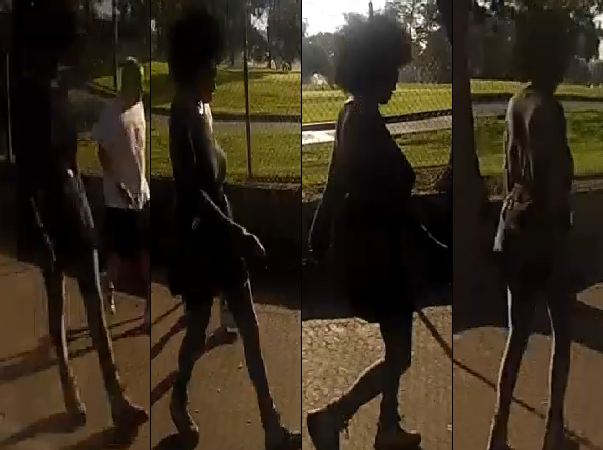}
        \caption{\textbf{The most ambiguous cases to label} for sex (far left), age (centre left), weight (centre right), and clothing style (far right), based on the entropy from the set of images labelled by 30 workers.}
        \label{fig-ambiguous_cases}
\end{figure*}
\section{Dataset Analysis}
\label{sec-analysis}
In this section, we analyse labelling error and explore the properties of the dataset. A summary of the dataset's statistics can be found in Table~\ref{tab-stats}. The distribution of bounding box widths and heights; the number of people per frame; and the distribution of track lengths can be found in Figure~\ref{fig-bbox_stats}. Occlusion statistics for pose keypoints; and a breakdown of the the number of labels for each class in each of the four subcategories can be found in Figure~\ref{fig-attribute_stats}.

Since the final bounding box estimates were derived from the pose labels and the tracks were verified by an oracle, the analysis of labelling error is focused on keypoints and fine-grained classes.

\subsection{Pose Error}
\label{sec-pose_error}
To estimate pose error, we take a keypoint from a particular sample (an image of a person). Since the location (x and y co-ordinates) of this keypoint was labelled by 3 different workers, the distance between each of the 3 locations can then be computed. The distance is normalised with respect to the height of the bounding box for the sample. The distribution of these distances across all samples, for each of the 14 keypoints can be found in Figure~\ref{fig-error}. 

The results indicate that workers tended to agree the most about the location of the chin. This is expected since the chin is a sharp, well defined, point on the face. The most disagreement occurred for the left shoulder and right hip. Both of these body parts are harder to localise since they are not sharp points. 

The keypoints are classified into 3 error classes - excellent (the green band) where error is less than 3\%, good (yellow band) where error is between 3-15\% and poor (red band) where error is greater than 15\%. The poor errors tend to occur when workers incorrectly exchange left and right labels. 

\subsection{Fine-Grained Label Error}
\label{sec-fg_error}
To estimate the error in the fine-grained labels, 180 tracks were randomly selected from the dataset and sent to MTURK to be labelled by 30 workers. The same process as outlined in Section~\ref{sec-attr} was followed except for the increase in annotators. 

Given a sample track, for each subcategory, the empirical probability distribution of its classes was calculated from the 30 worker labels. The normalised entropy for the sample was then computed using this distribution. A histogram of entropies for each subcategory across the 180 examples can be found in Figure~\ref{fig-error}. Low entropies indicate high agreement amongst annotators.

The fine-grained label error may be modelled statistically. Each one of the attributes we considered may be thought as varying in one dimension; thus, we assume that the $L$ class labels of a given attribute are the result of discretising a uniformly distributed continuous variable (e.g. age ranges from 0 to 100 years, and it is binned into five age categories).

We assume that each annotator may estimate the underlying continuous variable, with the addition of some `annotator noise' and produce a label by binning their continuous estimate. We model annotator noise as zero-mean with a free parameter $\sigma$, which we assume constant over the population of the annotators (a more sophisticated point of view may be found in~\cite{welinderBBP10}). In order to estimate $\sigma$ we fit this model to the empirical labelling results for each of the four subcategories (for convenience we rescaled the range of the underlying continuous variable to $(0,1)$).  Results are shown in Figure~\ref{fig-error}. The noise values that best fit the data are 0.07, 0.15, 0.11 and 0.23 for sex, age, weight and clothing respectively. This has to be compared to the size of the bins, which depends on the number of classes and is thus equal to 0.5, 0.2, 0.33 and 0.25 for each subcategory respectively. This means that the annotators' estimates of sex are excellent, quite consistent for weight, and somewhat vague for age and clothing, as one might expect.
Figure~\ref{fig-ambiguous_cases} contains examples of the 3 most ambiguous samples for each of the four subcategories.

\section{Baseline Experiments}
\label{sec-experiments}
In this section we present a set of baseline experiments for fine-grained categorisation and human pose estimation. The dataset is split into a training/validation set containing 4 videos, with the remaining 3 videos forming the test set. Since each video was collected on a unique day, different images of the same person \textbf{do not} appear in both the training and testing sets. While it is conceivable that a person appears in different videos while wearing the same clothes and in the same lighting conditions, we consider this to be unlikely.
\begin{table}[t]
\setlength{\tabcolsep}{1.5pt}
    \centering
        \begin{tabular}{@{} l *8c @{}}    \toprule
            \textbf{Parts} & Head & Torso & U.Arms & L.Arms & U.Legs & L.Legs & \textbf{Mean}  \\ \midrule
            \textbf{CRPa} & 40.6 & 59.7 & 26.1 & 13.6 & 31.2 & 21.5 & \textbf{32.1}\\
            \textbf{CRPb} & 66.2 & 88.1 & 60.7 & 30.2 & 71.9 & 60.9 & \textbf{63.0}\\
            \textbf{LSP} & 92.7 & 87.8 & 69.2 & 55.4 & 82.9 & 77.0 & \textbf{75.0}\\
        \bottomrule
        \end{tabular}
    \caption{\textbf{Pose estimation results.} We report the PCP for the parts in our dataset using the method described in Section~\ref{sec-pose_estimation}. CRPa corresponds to the pose model trained using the LSP dataset~\cite{Johnson2010} but tested on CRP. CRPb correponds to the case when the CRP dataset is used for both training and testing. Performance is best on the torso, with the lower arms and lower legs performing the most poorly. For comparison, results for a pose model trained and tested using the LSP dataset - the current pose estimation benchmark - are provided. The errors for our dataset are worse than those on the LSP dataset. This suggests that ours is a challenging dataset for pose estimation.}
    \label{tab-pose_res}
\end{table}
\subsection{Fine-Grained Categorisation}
\label{sec-fgc-res}
The fine-grained categorisation benchmark uses 'pose normalised deep convolutional nets' as proposed by Branson~\etal~\cite{Branson2014a}. In this framework, features are extracted by applying deep convolutional nets to image regions that are normalised by pose. It has state-of the-art performance on bird species categorisation and we believe that it will generalise to the people dataset.   
To elaborate, given a sample, image regions are extracted, then warped so that they are aligned to a set of prototypical models. Each warped region is then fed through a deep convolutional network where features are extracted from certain layers. The features from each warped region are then concatenated and used in a classifier. 

For this benchmark, given an image containing a person, two regions were extracted: 1) the full bounding box and 2) the portion of the bounding box that contains the shoulders, the hips and the head. Region (2) was then warped to a hand-defined prototype using a similarity alignment model, which was suggested by Branson~\etal to work best~\cite{Branson2014a}. The ground-truth keypoints for the left and right shoulder, the left and right hip, the top of the head, and the chin were used to compute the warping. 

The two regions were then fed through the pre-trained ImageNet convolutional neural network~\cite{Jia2014}. Features were extracted from the 5th layer after max-pooling (this layer gave the best performance). Features for each region were then concatenated and one-vs-all linear SVM's were used for classification. We refer to this method as ``bbox+body''. 

We also consider the case where only the bounding box is used as the extracted image region (no warping is applied). This method is referred to as ``bbox''.       

\begin{figure}[t]
    \centering
    \includegraphics[width=\linewidth]{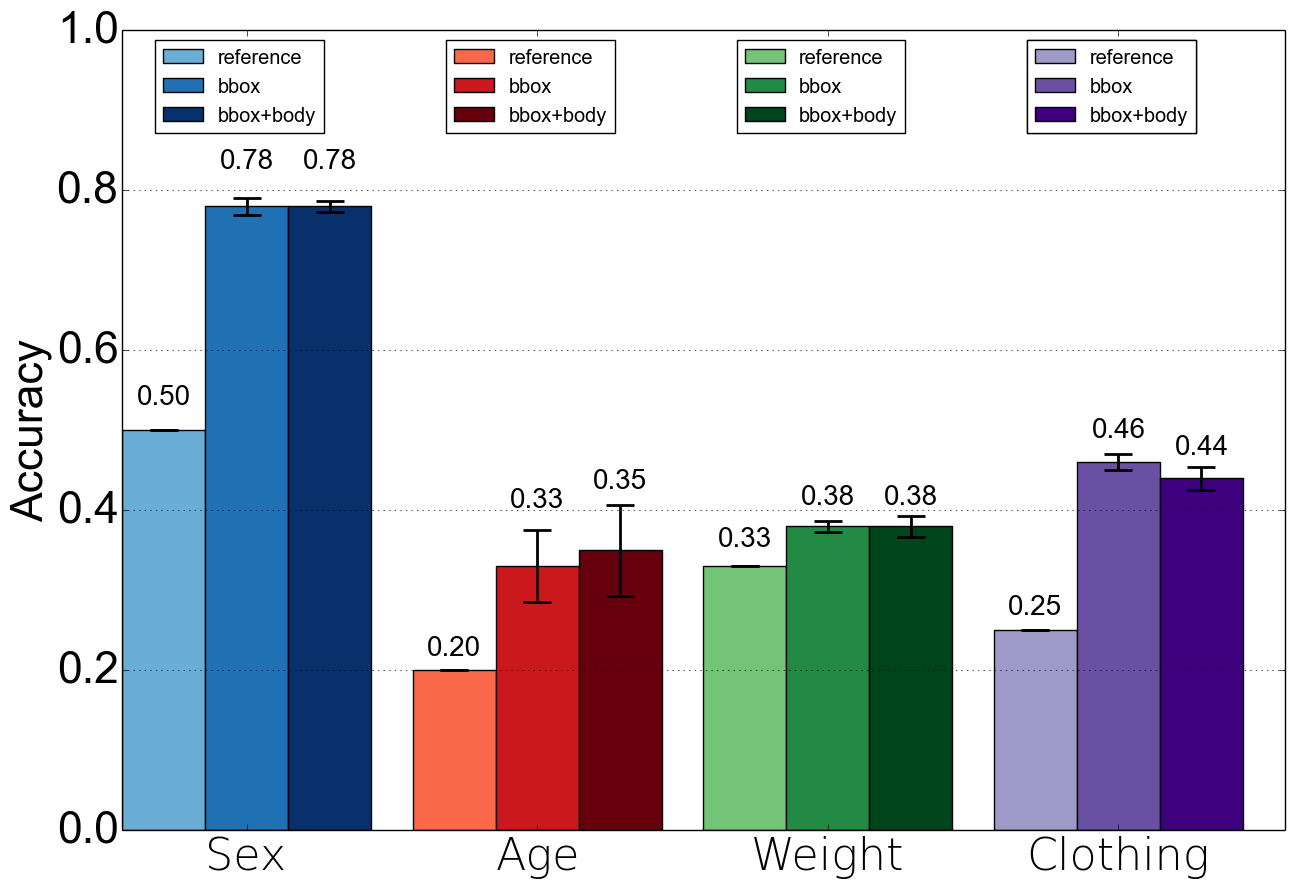}
    \caption{\textbf{Fine-grained classification results.} We report the mean average accuracy across 10 different train/test splits, for each of the subcategories, using the method of~\cite{Branson2014a}. Refer to Section~\ref{sec-fgc-res} for details. Average accuracy is computed assuming that there is a uniform prior across the classes. The reference value for each subcategory corresponds to chance. There is great room to improve classification accuracy for all subcategories. This suggests that this is a challenging dataset to work with.}
        \label{fig-fine_grained_results}
\end{figure}

Experiments were run for each of the subcategories in the dataset. We report the mean average classification accuracy across 10 trials. Each trial corresponded to a different train/test split. For all the experiments, keypoint occlusion information was not used and samples with missing parts were ignored. Training was done using code that was obtained directly from the authors~\cite{Branson2014a}. Results can be found in Figure~\ref{fig-fine_grained_results}. 

Aside from the sex subcategory, classification accuracy is low. This suggests that this is a challenging dataset. The results also show that for clothing style, using only the bounding box as the extracted image region is better than also including the pose aligned region; For weight and sex, very little difference is seen.  
\subsection{Pose Estimation}
\label{sec-pose_estimation}
Since most fine-grained categorisation techniques rely on parts, it is important to look at pose estimation. To benchmark human pose estimation we used the state-of-the-art, articulated pose estimator of Chen and Yuille~\cite{Chen2014}. This method extends Yang and Ramanan's work~\cite{Yang2011} to use deep features. The code is publicly available.

Two experiments were run using a single train/test split. In the first experiment, the pose model was trained using data from the LSP dataset~\cite{Johnson2010}. In the second, training was done using data from the CRP dataset. In both cases keypoint occlusion information was not used and samples with missing parts were ignored. 

The results are reported using a standard measure - the stricter interpretation of the percentage of correct parts (PCP)~\cite{Ramanan2006}. The results are shown in Table~\ref{tab-pose_res}

The results indicate that this dataset is more challenging than the existing pose estimation benchmark, the LSP dataset. The mean PCP is approximately 12\% lower than that of LSP. The high amount of occlusion present in our dataset is contributing to the poorer results.

\section{Discussion and Conclusions}
\label{sec-conclusions}
We introduce a video dataset designed to study fine-grained categorisation of people using the entire human body. Its novel and distinctive features are size, realism (natural behaviour, variety in viewpoint,  moving camera), fine-grained multi-label attributes (sex, weight, clothing, age), detailed annotations, and public availability.

Two sets of experiments were conducted to provide a performance baseline for the dataset. The first was a fine-grained categorisation task, where we used a state-of-the-art pose normalisation + deep network system~\cite{Branson2014a}. The second was a pose estimation task where we used an articulated pose model with deep features as the baseline method~\cite{Johnson2010}. We find that for both tasks the baseline method performance is low, and significantly lower than on other benchmark datasets. This suggests that our realistic and large dataset is challenging and will contribute to advance the state-of-the-art in both fine-grained classification and pose estimation.

A novel feature of our dataset is the occlusion labelling of the keypoints. Exploiting this information may be the first step towards improving performance for both tasks. Exploiting the temporal information is also worth exploring. Most pedestrians in our dataset appear multiple times over large intervals of time. We are planning on adding an identity label for each individual, to make our dataset useful for studying individual re-identification~\cite{Gray2008, Zheng2011} from a moving camera.
\section*{Acknowledgements}
This work is funded by the ARO-JPL NASA Stennis NAS7.03001 grant and the Gordon and Betty Moore Foundation.

{\small
\bibliographystyle{Style/ieee}
\bibliography{references}

\begin{thebibliography}{10}\itemsep=-1pt

\bibitem{Baluja2006}
S.~Baluja and H.~Rowley.
\newblock {Boosting Sex Identification Performance}.
\newblock {\em IJCV}, 71(1):111--119, June 2006.

\bibitem{Bashir2009}
K.~Bashir, T.~Xiang, and S.~Gong.
\newblock {Gait Recognition Using Gait Entropy Image}.
\newblock In {\em ICDP}, 2009.

\bibitem{Berg2014}
T.~Berg, J.~Liu, S.~W. Lee, M.~L. Alexander, D.~W. Jacobs, and P.~N. Belhumeur.
\newblock {Birdsnap: Large-Scale Fine-Grained Visual Categorization of Birds}.
\newblock In {\em CVPR}, 2014.

\bibitem{Bourdev2011}
L.~Bourdev, S.~Maji, and J.~Malik.
\newblock {Describing People: A Poselet-Based Approach to Attribute
  Classification}.
\newblock In {\em ICCV}, 2011.

\bibitem{Branson2014a}
S.~Branson, G.~{Van Horn}, P.~Perona, and S.~Belongie.
\newblock {Improved Bird Species Recognition Using Pose Normalized Deep
  Convolutional Nets}.
\newblock In {\em BMVC}, 2014.

\bibitem{Branson2010}
S.~Branson, C.~Wah, F.~Schroff, B.~Babenko, P.~Perona, and S.~Belongie.
\newblock {Visual Recognition with Humans in the Loop}.
\newblock In {\em ECCV}, 2010.

\bibitem{Cao2008}
L.~Cao, M.~Dikmen, Y.~Fu, and T.~S. Huang.
\newblock {Gender recognition from body}.
\newblock In {\em ACM}, 2008.

\bibitem{Chen2014}
X.~Chen and A.~Yuille.
\newblock {Articulated Pose Estimation by a Graphical Model with Image
  Dependent Pairwise Relations}.
\newblock In {\em NIPS}, 2014.

\bibitem{Collins2009}
M.~Collins, J.~Zhang, and P.~Miller.
\newblock {Full body image feature representations for gender profiling}.
\newblock In {\em ICCV Workshop}, Sept. 2009.

\bibitem{Cottrell1990}
G.~Cottrell and J.~Metcalfe.
\newblock {EMPATH: Face, Emotion, and Gender Recognition using Holons}.
\newblock In {\em NIPS}, 1990.

\bibitem{Deng2009}
J.~Deng, W.~Dong, R.~Socher, L.~Li-Jia, K.~Li, and L.~Fei-Fei.
\newblock {ImageNet: A Large-Scale Hierarchical Image Database}.
\newblock In {\em CVPR}, 2009.

\bibitem{Dollar2009a}
P.~Dollar, C.~Wojek, B.~Schiele, and P.~Perona.
\newblock {Pedestrian detection: A benchmark}.
\newblock In {\em CVPR}, 2009.

\bibitem{Fu2014}
S.~Fu, H.~He, and Z.~Hou.
\newblock {Learning Race from Face: A Survey}.
\newblock {\em PAMI}, 99(PP):1, 2014.

\bibitem{Fu2010}
Y.~Fu, G.~Guo, and T.~S. Huang.
\newblock {Age synthesis and estimation via faces: a survey.}
\newblock {\em PAMI}, 32(11):1955--76, Nov. 2010.

\bibitem{Golomb1990}
B.~Golomb, D.~Lawrence, and T.~Sejnowski.
\newblock {SEXNET: A Neural Network Identifies Sex From Human Faces.}
\newblock In {\em NIPS}, 1990.

\bibitem{Gray2008}
D.~Gray and H.~Tao.
\newblock {Viewpoint Invariant Pedestrian Recognition with an Ensemble of
  Localized Features}.
\newblock In {\em ECCV}, 2008.

\bibitem{Han2006}
J.~Han and B.~Bhanu.
\newblock {Individual Recognition Using Gait Energy Image.}
\newblock {\em PAMI}, 28(2):316--22, Mar. 2006.

\bibitem{Iwama2012}
H.~Iwama, M.~Okumura, Y.~Makihara, and Y.~Yagi.
\newblock {The OU-ISIR Gait Database Comprising the Large Population Dataset
  and Performance Evaluation of Gait Recognition}.
\newblock {\em IEEE Trans. Inf. Forensics Security}, 7(5):1511--1521, Oct.
  2012.

\bibitem{Jia2014}
Y.~Jia, E.~Shelhamer, J.~Donahue, S.~Karayev, J.~Long, R.~Girshick,
  S.~Guadarrama, and T.~Darrell.
\newblock Caffe: Convolutional architecture for fast feature embedding.
\newblock Technical report, 2014.

\bibitem{Johnson2010}
S.~Johnson and M.~Everingham.
\newblock {Clustered Pose and Nonlinear Appearance Models for Human Pose
  Estimation}.
\newblock In {\em BMVC}, 2010.

\bibitem{Khosla2011}
A.~Khosla, N.~Jayadevaprakash, B.~Yao, and L.~Fei-Fei.
\newblock {Novel Dataset for Fine-Grained Image Categorization: Stanford Dogs}.
\newblock In {\em FGVC Workshop, CVPR}, 2011.

\bibitem{Kumar2008}
N.~Kumar, P.~Belhumeur, and S.~Nayar.
\newblock {FaceTracer: A Search Engine for Large Collections of Images with
  Faces}.
\newblock {\em ECCV}, 2008.

\bibitem{Kumar2012}
N.~Kumar, P.~N. Belhumeur, A.~Biswas, D.~W. Jacobs, W.~J. Kress, I.~Lopez, and
  V.~B. Soares.
\newblock {Leafsnap: A Computer Vision System for Automatic Plant Species
  Identification}.
\newblock In {\em ECCV}, 2012.

\bibitem{Lin2014}
T.-Y. Lin, M.~Maire, S.~Belongie, J.~Hays, P.~Perona, D.~Ramanan, P.~Dollár,
  and C.~Zitnick.
\newblock Microsoft coco: Common objects in context.
\newblock In {\em ECCV}, 2014.

\bibitem{Liu2012a}
J.~Liu, A.~Kanazawa, D.~Jacobs, and P.~Belhumeur.
\newblock {Dog Breed Classification Using Part Localization}.
\newblock In {\em ECCV}, 2012.

\bibitem{Maji2013}
S.~Maji, J.~Kannala, E.~Rahtu, A.~Vedaldi, and M.~Blaschko.
\newblock {Fine-Grained Visual Classification of Aircraft}.
\newblock Technical report, 2013.

\bibitem{Makihara2012}
Y.~Makihara, H.~Mannami, A.~Tsuji, M.~A. Hossain, K.~Sugiura, A.~Mori, and
  Y.~Yagi.
\newblock {The OU-ISIR Gait Database Comprising the Treadmill Dataset}.
\newblock {\em IPSJ Transactions on Computer Vision and Applications},
  4:53--62, 2012.

\bibitem{Mart2009}
G.~Mart, D.~Lytle, A.~Moldenke, and E.~Mortensen.
\newblock {Dictionary-Free Categorization of Very Similar Objects via Stacked
  Evidence Trees}.
\newblock In {\em CVPR}, 2009.

\bibitem{Moghaddam2002}
B.~Moghaddam and M.-H. Yang.
\newblock {Learning Gender With Support Faces}.
\newblock {\em PAMI}, 24(5):707--711, May 2002.

\bibitem{Nilsback2006}
M.-E. Nilsback and A.~Zisserman.
\newblock {A Visual Vocabulary for Flower Classification}.
\newblock In {\em CVPR}, 2006.

\bibitem{Nilsback2008}
M.-E. Nilsback and A.~Zisserman.
\newblock {Automated Flower Classification over a Large Number of Classes}.
\newblock In {\em ICCVGIP}, 2008.

\bibitem{Oren1997}
M.~Oren, C.~Papageorgiou, P.~Sinha, E.~Osuna, and T.~Poggio.
\newblock {Pedestrian detection using wavelet templates}.
\newblock In {\em CVPR}, 1997.

\bibitem{Parkhi2012}
O.~M. Parkhi, A.~Vedaldi, A.~Zisserman, and C.~V. Jawahar.
\newblock {Cats and Dogs}.
\newblock In {\em CVPR}, 2012.

\bibitem{Ramanan2006}
D.~Ramanan.
\newblock {Learning to Parse Images of Articulated Bodies}.
\newblock In {\em NIPS}, 2006.

\bibitem{Sarkar2005}
S.~Sarkar, P.~J. Phillips, Z.~Liu, I.~R. Vega, P.~Grother, and K.~W. Bowyer.
\newblock {The HumanID Gait Challenge Problem: Data Sets, Performance, and
  Analysis.}
\newblock {\em PAMI}, 27(2):162--77, Mar. 2005.

\bibitem{Shakhnarovich2002}
G.~Shakhnarovich, P.~Viola, and B.~Moghaddam.
\newblock {A Unified Learning Framework for Real Time Face Detection and
  Classification}.
\newblock In {\em Automatic Face and Gesture Recognition}, 2002.

\bibitem{Stark2012}
M.~Stark, J.~Krause, B.~Pepik, D.~Meger, J.~Little, B.~Schiele, and D.~Koller.
\newblock {Fine-Grained Categorization for 3D Scene Understanding}.
\newblock In {\em BMVC}, 2012.

\bibitem{Wah2011}
C.~Wah, S.~Branson, P.~Welinder, P.~Perona, and S.~Belongie.
\newblock {The Caltech-UCSD Birds-200-2011 Dataset}.
\newblock Technical report, California Institute of Technology, 2011.

\bibitem{Wang2009}
J.~Wang, K.~Markert, and M.~Everingham.
\newblock {Learning Models for Object Recognition from Natural Language
  Descriptions}.
\newblock In {\em BMVC}, 2009.

\bibitem{welinderBBP10}
P.~Welinder, S.~Branson, S.~Belongie, and P.~Perona.
\newblock The multidimensional wisdom of crowds.
\newblock In {\em NIPS}, 2010.

\bibitem{Yang2011}
Y.~Yang and D.~Ramanan.
\newblock {Articulated Pose Estimation using Flexible Mixtures of Parts}.
\newblock In {\em CVPR}, 2011.

\bibitem{Zhang2014}
N.~Zhang, M.~Paluri, M.~A. Ranzato, T.~Darrell, and L.~Bourdev.
\newblock {PANDA : Pose Aligned Networks for Deep Attribute Modeling}.
\newblock In {\em CVPR}, 2014.

\bibitem{Zheng2011}
W.~Zheng, S.~Gong, and T.~Xiang.
\newblock {Person Re-identification by Probabilistic Relative Distance
  Comparison}.
\newblock In {\em CVPR}, 2011.

\end{thebibliography}
}

\end{document}